\documentclass{bmvc2k}
\usepackage{booktabs}
\usepackage{amsmath}
\usepackage{array}
\usepackage[normalem]{ulem} 
\usepackage{algorithm}
\usepackage{algpseudocode}
\usepackage[table]{xcolor}
\usepackage{wrapfig}
\graphicspath{{figures/}}

\title{FiRe: Fixed-Noise Refinement \\for Visual Counterfactual Explanations}

\addauthor{Yan Zeng}{s242652@dtu.dk}{1}
\addauthor{Changlu Guo}{chagu@dtu.dk}{1}
\addauthor{Oskar Kristoffersen}{ofhkr@dtu.dk}{1}
\addauthor{Anders Nymark Christensen}{anym@dtu.dk}{1}
\addauthor{Morten Rieger Hannemose}{mohan@dtu.dk}{1}
\addauthor{Anders Bjorholm Dahl}{abda@dtu.dk}{1}

\addinstitution{
Department of Applied Mathematics and Computer Science, Technical University of Denmark
}

\runninghead{Zeng et al.}{FiRe}

\begin{document}

\maketitle

\begin{abstract}
Visual counterfactual explanations aim to change classifier decisions through realistic and localized edits while preserving decision-irrelevant content. Existing DDPM-based methods typically perform classifier-guided editing along a long reverse denoising trajectory. The changing noise levels make semantic editability and spatial control difficult to balance, and the editable state is noisy, whereas the target classifier is trained on clean images. As a result, these methods require either costly recursive denoising or low-quality one-step estimates to obtain classifier-facing clean images. We propose FiRe, a Fixed-noise Refinement framework for visual counterfactual explanations. Rather than following a reverse denoising trajectory, FiRe maps the input to a fixed noise level and iteratively refines the noisy state at that level. To provide clean images for classifier guidance, FiRe first adapts Pixel Mean Flow to visual counterfactual explanation, enabling direct clean-image prediction from noisy states. To make fixed-noise refinement produce minimal and localized counterfactual edits, FiRe introduces three FiRe-specific controls: a dynamic dual-mask strategy, adaptive guidance, and early stopping, which determine where edits accumulate, which changes become visible, and when refinement stops. Experiments on five tasks across three datasets show that, compared with the strongest recent baseline, FiRe achieves about 3$\times$ faster online inference and 8$\times$ fewer FLOPs while obtaining comparable or state-of-the-art counterfactual quality. The project page is available at \href{https://yan-zen9.github.io/FiRe/}{https://yan-zen9.github.io/FiRe/}.
\end{abstract}

\section{Introduction}
\label{sec:intro}

\begin{figure}[t]
    \centering
    \includegraphics[width=0.9\linewidth]{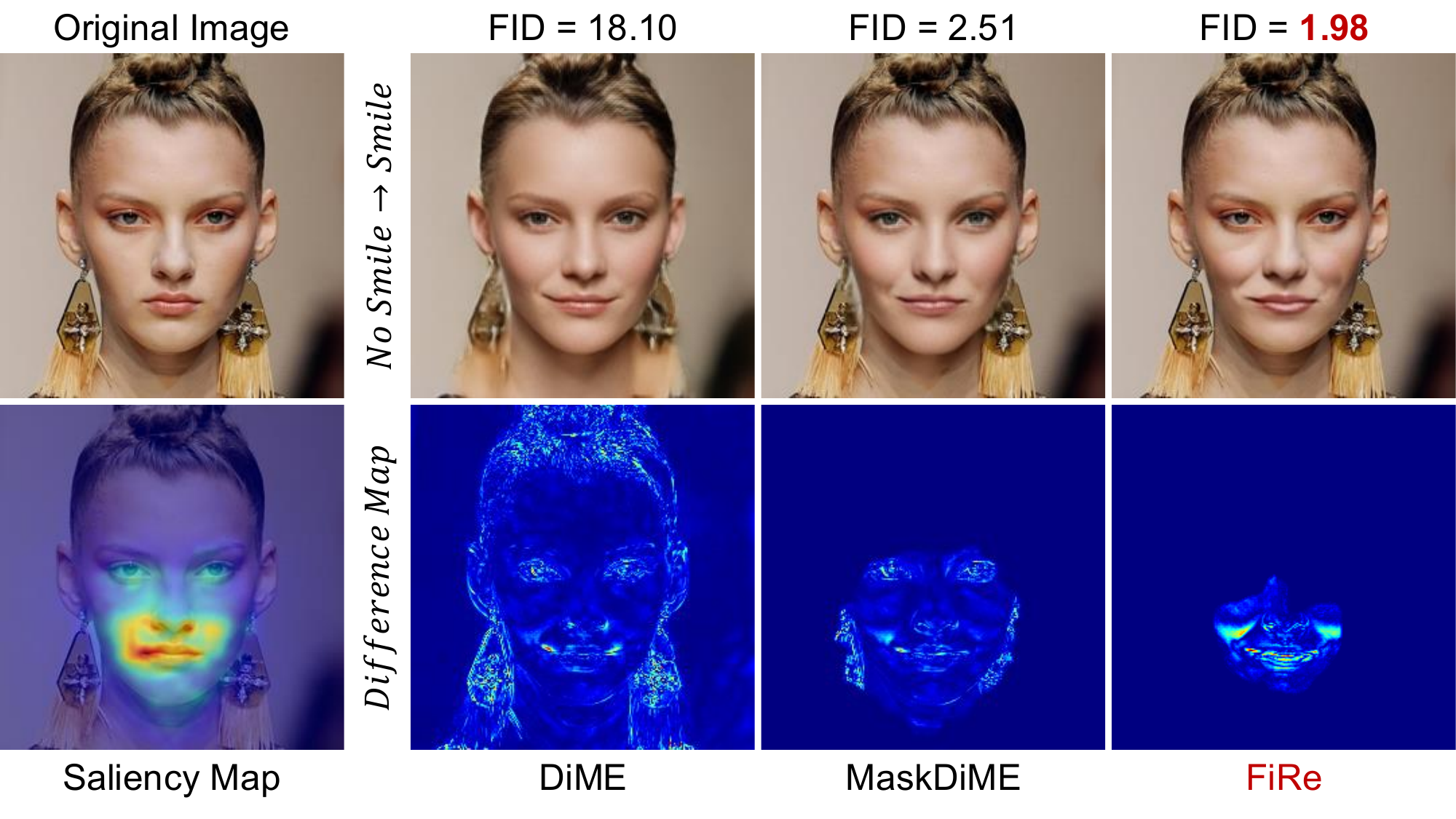}
\caption{
\textbf{Compared with DiME and MaskDiME,} FiRe produces a cleaner and more localized edit that better matches the classifier-relevant region.
}
    \label{fig:diff}
\end{figure}

Recent visual counterfactual explanation methods~\cite{goyal2019counterfactual,jacob2022steex,rodriguez2021beyond} often use Denoising Diffusion Probabilistic Models (DDPMs)~\cite{ho2020denoising,dhariwal2021diffusion} because they provide strong image priors. However, most DDPM-based methods~\cite{jeanneret2022diffusion,augustin2022diffusion,weng2024fast,guo2026maskdime} perform classifier-guided editing along a reverse denoising trajectory~\cite{ho2020denoising}, so counterfactual optimization must proceed through a sequence of changing noise levels. This introduces a timestep-dependent conflict between semantic editability and spatial control. At early high-noise stages, the model has a larger editing space, but the resulting changes are difficult to localize and may affect irrelevant regions. At later low-noise stages, the state is closer to the clean image domain, but the editable space has become more restricted and may no longer support the required semantic change.

Another key difficulty is the mismatch between the noisy editable state and the clean-image input required by the classifier. Since the target classifier is trained on clean images, DDPM-based methods must construct a classifier-facing clean image from the noisy state at each guidance step. One line of methods obtains a more reliable clean image through recursive denoising~\cite{jeanneret2022diffusion,augustin2022diffusion}, but at a high computational cost; another line uses a one-step clean-image estimate~\cite{weng2024fast,guo2026maskdime}, which is more efficient but produces lower-quality images. As a result, these two types of methods are respectively limited by speed and quality bottlenecks.

To address these limitations, we introduce FiRe, reformulating visual counterfactual explanation as fixed-noise refinement. Instead of editing along a DDPM reverse trajectory with changing noise levels, FiRe maps the input image to an intermediate noise level and iteratively refines the noisy state at this level. This design preserves sufficient semantic editability for the target change while reducing the uncontrolled global changes caused by high-noise editing. To obtain high-quality classifier-facing clean images efficiently, FiRe uses Pixel Mean Flow (PMF)~\cite{lu2026one} as a clean-image predictor. PMF directly predicts a clean image from a noisy state in one step, avoiding both recursive denoising and low-quality one-step estimates. In this way, FiRe addresses the timestep-varying editability of reverse-trajectory editing and alleviates the speed and quality bottlenecks of existing DDPM-based methods.

While fixed-noise refinement and PMF address the trajectory and clean-image prediction issues, visual counterfactual explanations still require localized and minimal edits. This requires a spatial control mechanism designed for the fixed-noise setting. Existing DDPM-based masks~\cite{weng2024fast,guo2026maskdime} operate along changing noise levels, where each timestep is usually visited only once and the mask controls the current update. In FiRe, all updates occur at the same fixed noise level, so counterfactual changes are accumulated through multiple refinement steps in the same noisy state. Edit regions discovered early should therefore remain active later; otherwise, partially formed counterfactual changes may be weakened or interrupted. FiRe further separates noisy-state control from clean-image visibility control. The noisy state is the optimization space and should keep sufficient editable regions for accumulating target-related changes, whereas the clean image is both the classifier-facing input and the final output, whose visible changes should be more tightly constrained. FiRe therefore uses one mask to control where changes accumulate in the noisy state and another to control which changes become visible in the clean image. This preserves edit continuity while limiting unnecessary visible modifications.

Even with dynamic dual-mask spatial control, FiRe still needs to regulate the guidance strength. Since all updates act on the same fixed noisy state, a fixed guidance strength can either over-edit easy samples after the target decision has already been reached, or be too weak for hard samples to cross the decision boundary. Fixed-noise refinement therefore requires update strength to adapt to both refinement progress and target confidence. FiRe uses adaptive guidance and early stopping: the guidance is strengthened when more editing is needed, weakened as the target decision is approached, and stopped once the counterfactual is sufficient to change the classifier decision. This prevents redundant changes from accumulating in the fixed noisy state and leads to more localized counterfactual edits.

The main contributions of this work are summarized as follows:
\begin{enumerate}
    \item We present FiRe, a framework that reformulates classifier-guided visual counterfactual explanation as iterative refinement at a single intermediate noise level, avoiding the timestep-dependent limitations of DDPM reverse trajectories.

    \item FiRe enables direct classifier-facing clean-image prediction from noisy states without recursive denoising or low-quality one-step estimates.

    \item  We design controls tailored to fixed-noise refinement, including a dynamic dual-mask strategy, adaptive guidance, and early stopping, to preserve edit continuity and reduce unnecessary visible modifications, as shown in Figure~\ref{fig:diff}.

    \item We evaluate FiRe on five tasks across CelebA, CelebA-HQ, and CheXpert. Compared with the strongest recent baseline, FiRe achieves about 3$\times$ faster inference and 8$\times$ fewer FLOPs, with comparable or state-of-the-art counterfactual quality.
\end{enumerate}

\section{Related Work}

\subsection{Explainable Artificial Intelligence and Visual Counterfactuals}

Explainable artificial intelligence (XAI)~\cite{arrieta2020explainable} aims to make model decisions more understandable. 
A common setting is post-hoc explanation, where a pretrained model is explained without modifying its architecture. 
Representative methods include saliency maps~\cite{simonyan2013deep,smilkov2017smoothgrad,selvaraju2017grad}, concept attribution~\cite{kim2018interpretability}, model distillation~\cite{hendricks2016generating}, and counterfactual explanations~\cite{wachter2017counterfactual,goyal2019counterfactual}. 
This work focuses on visual counterfactual explanations (VCEs)~\cite{goyal2019counterfactual,rodriguez2021beyond,jacob2022steex}, which generate alternative images that change a classifier's prediction while preserving decision-irrelevant content. 
In contrast to saliency maps and concept attribution, which identify regions or concepts associated with a prediction, VCEs explain a decision by showing what visual evidence would need to change for the classifier to predict a different class. Sparse image-space VCEs~\cite{boreiko2022sparse} optimize a sparse perturbation model with Auto-Frank-Wolfe to produce localized counterfactuals without relying on a generative reverse trajectory.

\subsection{Diffusion-Based Visual Counterfactual Explanations}

Visual counterfactual explanations have been studied with several classes of generative models, including VAEs~\cite{joshi2018xgems}, GANs~\cite{lang2021explaining,choi2018stargan,he2019attgan}, and diffusion models~\cite{jeanneret2022diffusion,augustin2022diffusion,weng2024fast,guo2026maskdime}. Among them, diffusion-based methods have become a major direction because of their strong image priors. DiME~\cite{jeanneret2022diffusion} first introduced DDPM reverse denoising trajectories~\cite{ho2020denoising} into visual counterfactual explanation by guiding the reverse process with target-class classifier gradients, which allows the method to produce realistic counterfactual images. This formulation, however, is computationally expensive, since obtaining classifier-facing clean images from noisy states usually requires recursive denoising. ACE~\cite{jeanneret2023adversarial} instead starts from adversarial optimization and uses a diffusion prior to regularize the perturbation, making the resulting changes more semantic. Its main effect is to improve semantic plausibility, rather than to reduce the high computational cost of DiME. FastDiME~\cite{weng2024fast} addresses this efficiency bottleneck by using Tweedie's formula~\cite{robbins1992empirical} to obtain a one-step clean-image estimate from the DDPM noise prediction. This reduces computation, but the estimated clean image may introduce approximation errors, which can lead to unstable classifier gradients, and spatial locality remains limited. RCSB~\cite{sobieski2025rethinking} improves locality by restricting edits to selected regions, but fixed-region constraints may not adapt well to the changing semantic evidence during counterfactual generation. MaskDiME~\cite{guo2026maskdime} further addresses this issue by constructing adaptive masks from classifier gradients, allowing the edited region to change with the classifier evidence.

Despite these improvements in efficiency and locality, these methods still operate within the DDPM reverse denoising trajectory~\cite{jeanneret2022diffusion,augustin2022diffusion}, where changing noise levels make semantic editability and spatial control difficult to balance. They also depend on either costly recursive denoising~\cite{jeanneret2022diffusion,augustin2022diffusion} or lower-quality one-step approximations~\cite{weng2024fast,guo2026maskdime} to obtain the clean images required by the classifier. FiRe addresses this shared limitation by replacing reverse-trajectory editing with fixed-noise refinement and using Pixel Mean Flow (PMF)~\cite{lu2026one} to directly predict classifier-facing clean images.

Other recent visual counterfactual explanation methods build on latent diffusion models~\cite{rombach2022high}, such as LDCE~\cite{farid2023latent}; use text guidance, such as TiME~\cite{jeanneret2024text} and PRISM~\cite{kumar2025prism}; or introduce additional supervision, such as DeCoDEx~\cite{fathi2024decodex}. These methods are relevant to visual counterfactual explanation, but they rely on additional conditioning or supervision beyond the target classifier. Therefore, they are not directly comparable baselines in this work.

\subsection{Clean-Image Prediction and Pixel Mean Flow}

Flow Matching~\cite{lipman2022flow} provides a generative modeling framework that differs from the DDPM reverse process by learning a vector field between noise and data distributions. Rectified Flow~\cite{liu2023flow,liu2022rectified} further straightens the transport trajectories, enabling accurate generation with fewer integration steps. InstaFlow~\cite{liu2024instaflow} demonstrates one-step generation, while rectified-flow priors~\cite{yang2025rectified} extend this family to image inversion and editing. Although these methods reduce sampling cost or provide efficient editing priors, they are not designed for classifier-guided VCE and do not specify how to change a classifier decision, localize the resulting edit, or determine when refinement should stop.

Pixel Mean Flow (PMF)~\cite{lu2026one} directly predicts a clean image from a noisy input in one step. This property is particularly suitable for classifier-guided VCE, where the classifier requires a clean image at every refinement step to compute reliable guidance. PMF itself remains a general image generation model rather than a counterfactual explanation method. FiRe integrates PMF clean-image prediction into fixed-noise refinement and introduces dynamic dual masking, adaptive guidance, and early stopping to produce localized and target-valid counterfactual explanations.


\section{Method}
\label{sec:method}

\begin{figure}[t]
    \centering
    \includegraphics[width=0.9\linewidth]{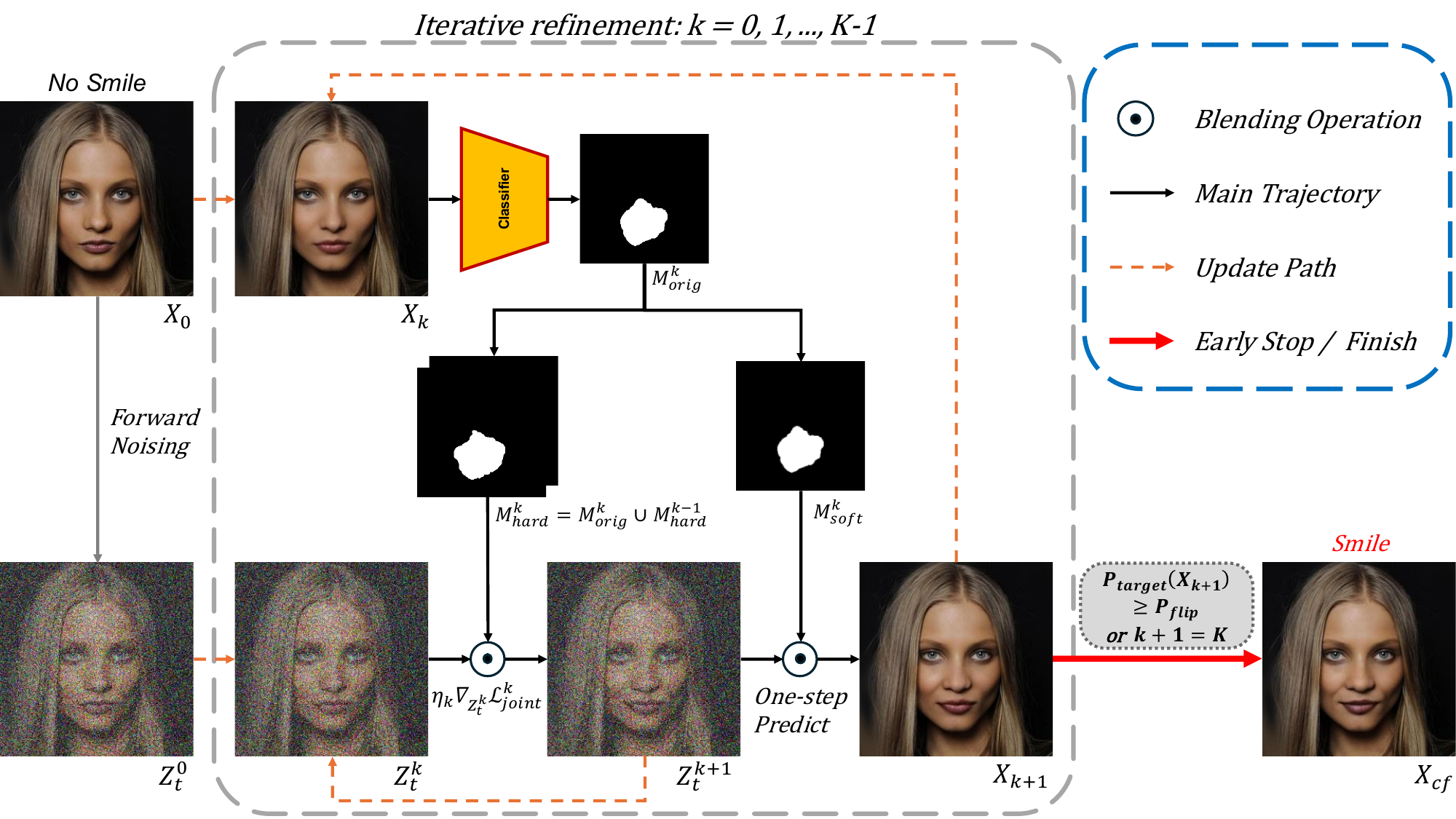}
\caption{
\textbf{Overview of FiRe.}
Given an input image $X_0$, FiRe first applies forward noising to obtain a fixed noisy state $Z_t^0$, then performs iterative refinement at the same noise level. At step $k$, the current clean image $X_k$ is used to compute a classifier-driven original mask $M_{\mathrm{orig}}^k$ and adaptive gradient guidance $\eta_k\nabla_{Z_t^k}\mathcal{L}_{\mathrm{joint}}^k$. FiRe derives two masks from $M_{\mathrm{orig}}^k$: a union hard mask $M_{\mathrm{hard}}^k$, which keeps previously selected edit regions active and constrains the noisy-state update from $Z_t^k$ to $Z_t^{k+1}$, and a soft mask $M_{\mathrm{soft}}^k$, which constrains the clean-image update from the updated noisy state $Z_t^{k+1}$ to the next clean image $X_{k+1}$. Refinement stops once $p_{\mathrm{target}}(X_{k+1}) \ge P_{\mathrm{flip}}$, or after $K$ updates.
}
    \label{fig:arc}
\end{figure}

\subsection{Overview}
Figure~\ref{fig:arc} shows the FiRe pipeline.
Given an input image $X_0$, FiRe first maps it to a fixed noisy state $Z_t^0$ and then applies $K$ refinement updates at the same noise level.
At each step, PMF predicts a clean classifier-facing image.
The classifier then provides attribution for spatial control and gradients for counterfactual guidance.
Here, the refinement index $k$ denotes an optimization step at fixed $t$, rather than a DDPM timestep.
This fixed-noise formulation requires two types of control: spatial control to keep repeated edits localized, and guidance control to stop unnecessary updates once the target decision has been reached.

\subsection{Problem Setup}
\label{sec:problem_setup}

Let $X_0 \in [-1,1]^{C\times H\times W}$ be the input image and $f$ be the fixed classifier to be explained. We denote the target label as $y_{\mathrm{target}}$. The goal is to generate a counterfactual image $X_{\mathrm{cf}}$ that changes the classifier prediction toward $y_{\mathrm{target}}$ while preserving decision-irrelevant content from $X_0$. The target probability is defined as
\begin{equation}
p_{\mathrm{target}}(X)
= \mathrm{softmax}(f(X))_{y_{\mathrm{target}}}.
\label{eq:target_prob}
\end{equation}

\subsection{Fixed-Noise State Initialization}
\label{sec:fixed_noise_initialization}

FiRe first perturbs the input image to a fixed noise level~\cite{lu2026one}:
\begin{equation}
Z_t^0 = (1-t)X_0 + tZ^0,
\qquad
Z^0 \sim \mathcal{N}(0,I).
\label{eq:fixed_noise_init}
\end{equation}
The noise level $t$ is kept fixed throughout refinement. Thus, $k$ indexes refinement updates rather than DDPM timesteps, and the sequence $Z_t^0, Z_t^1, \ldots, Z_t^K$ is not a reverse denoising trajectory. Each update modifies the noisy state at the same noise level, and a clean image is predicted from the updated noisy state for classifier guidance and output. We use $X_0$ as the initial clean image for constructing the first original mask.

\subsection{Classifier-Driven Dynamic Dual Mask}
Given the current clean image $X_k$, FiRe uses the target classifier to locate the region most relevant to the decision.
Let $s_{y_{\mathrm{target}}}(f(X))$ denote the target-class logit.
We compute the attribution heatmap as
\begin{equation}
A^k =
\left.
\left|
\nabla_X s_{y_{\mathrm{target}}}(f(X))
\right|
\right|_{X=X_k}.
\end{equation}
In practice, we use SmoothGrad~\cite{smilkov2017smoothgrad} to stabilize $A^k$ by averaging gradients over $N$ perturbed copies of $X_k$ in parallel.
FiRe then selects the top-$\rho$ salient region, keeps the largest connected component, and dilates it to obtain the current binary original mask:
\begin{equation}
C^k = \mathrm{LCC}(\mathrm{Top}_{\rho}(A^k)),
\quad
M_{\mathrm{orig}}^k = \mathrm{Dilate}(C^k; r).
\label{eq:orig_mask}
\end{equation}
Here, $\rho$ denotes the saliency ratio, so $\mathrm{Top}_{\rho}(A^k)$ selects the top $\rho$ fraction of pixels with the largest attribution values, and $r$ is the dilation radius. 

The fixed-noise setting calls for a mask design different from MaskDiME~\cite{guo2026maskdime}.
MaskDiME builds dynamic masks along a DDPM reverse denoising trajectory, where both the editable state and noise level change at each timestep.
FiRe instead performs all refinement steps at the same noise level.
If the update uses only the current mask $M_{\mathrm{orig}}^k$, regions selected in earlier steps may be removed after mask recomputation, weakening or interrupting accumulated counterfactual changes.
To avoid this, FiRe uses a union hard mask as edit memory:
\begin{equation}
M_{\mathrm{hard}}^k = M_{\mathrm{hard}}^{k-1} \cup M_{\mathrm{orig}}^k,
\quad
M_{\mathrm{hard}}^{-1}=0.
\label{eq:hard_mask}
\end{equation}
This mask is applied only to the noisy-state update, so previously selected regions remain editable throughout refinement.
The binary mask $M_{\mathrm{orig}}^k$ is not directly suitable for the visible clean image, since hard boundaries can introduce artifacts after blending.
FiRe therefore constructs a feathered soft mask:
\begin{equation}
M_{\mathrm{soft}}^k = \mathrm{Feather}(M_{\mathrm{orig}}^k; w),
\label{eq:soft_mask}
\end{equation}
where $w$ controls the smooth transition around the mask boundary.
The soft mask is used only for clean-image blending and final visible modification control.
In this design, the hard mask determines where counterfactual edits accumulate in the noisy state, while the soft mask determines which generated changes are visible in the classifier-facing clean image.

\subsection{Adaptive Guidance for Fixed-Noise Refinement}
\label{sec:masked_guidance_update}

FiRe computes classifier guidance on the current clean image $X_k$. The objective combines target classification with content-preserving regularization:
\begin{equation}
\mathcal{L}_{\mathrm{joint}}^k =
\lambda_{\mathrm{cls}}\mathcal{L}_{\mathrm{cls}}(X_k,y_{\mathrm{target}})
+ \lambda_{\mathrm{perc}}\mathcal{L}_{\mathrm{perc}}(X_k,X_0)
+ \lambda_{\mathrm{tv}}\mathcal{L}_{\mathrm{tv}}(X_k,X_0).
\label{eq:joint_loss}
\end{equation}
The classification loss moves the counterfactual toward the target class, while the perceptual~\cite{johnson2016perceptual,zhang2018unreasonable} and TV losses~\cite{rudin1992nonlinear} discourage changes to decision-irrelevant content and suppress local artifacts.

The loss is evaluated on the clean image, but the update is applied to the noisy state. FiRe obtains the noisy-state gradient by backpropagating the clean-image gradient through the PMF~\cite{lu2026one} prediction and clean-image update path:
\begin{equation}
\nabla_{Z_t^k}\mathcal{L}_{\mathrm{joint}}^k
=
\left(\frac{\partial X_k}{\partial Z_t^k}\right)^{\top}
\nabla_{X_k}\mathcal{L}_{\mathrm{joint}}^k .
\label{eq:clean_to_noisy_gradient}
\end{equation}
Here, $\partial X_k / \partial Z_t^k$ follows the computation path that forms the current clean image from the current noisy state, including PMF prediction and soft-mask clean-image update. We normalize $\nabla_{Z_t^k}\mathcal{L}_{\mathrm{joint}}^k$ before applying it to the noisy-state update, and use the same notation below for simplicity.

A fixed guidance strength is not well suited to fixed-noise refinement. 
Since FiRe repeatedly updates the same noisy state at one noise level, a large constant step can over-edit easy samples after they are already close to the target decision, while a small constant step may fail on harder samples. 
We therefore use a confidence-progress adaptive guidance scale:
\begin{equation}
\eta_k =
\begin{cases}
0.3\eta, & k = 0,\\
\eta \frac{k}{K-1}
\left[
\frac{P_{\mathrm{flip}} - p_{\mathrm{target}}(X_k)}
{P_{\mathrm{flip}}}
\right]_+, & k > 0,
\end{cases}
\label{eq:adaptive_step}
\end{equation}
where $\eta$ is the base guidance scale, $K$ is the maximum number of refinement updates, $P_{\mathrm{flip}}$ is the target-probability threshold, and $[\cdot]_+$ clamps negative values to zero. 
The progress term increases guidance as refinement proceeds, while the confidence term suppresses guidance as the target probability approaches the stopping threshold. 
This makes the update stronger only when needed and reduces unnecessary refinement after the counterfactual becomes sufficient. The adaptive update is applied under the union hard mask:
\begin{equation}
Z_t^{k+1}
=
M_{\mathrm{hard}}^k \odot
\left(
Z_t^k - \eta_k \nabla_{Z_t^k}\mathcal{L}_{\mathrm{joint}}^k
\right)
+
(1-M_{\mathrm{hard}}^k)\odot Z_t^0.
\label{eq:masked_update}
\end{equation}
Inside $M_{\mathrm{hard}}^k$, FiRe accumulates counterfactual changes. 
Outside the mask, the noisy state is reset to the initial noisy reference $Z_t^0$, preventing drift in decision-irrelevant regions.

\begin{algorithm}[t]
\caption{FiRe: Fixed-noise Counterfactual Refinement}
\label{alg:fire}
\begin{algorithmic}[1]
\Require $X_0$, classifier $f$, target label $y_{\mathrm{target}}$, PMF predictor $G_\theta$, noise level $t$, update budget $K$, threshold $P_{\mathrm{flip}}$
\Ensure Counterfactual image $X_{\mathrm{cf}}$
\State Sample $Z^0\sim\mathcal{N}(0,I)$ and initialize $Z_t^0\gets (1-t)X_0+tZ^0$
\State Initialize $M_{\mathrm{hard}}^{-1}\gets\mathbf{0}$ and set $X_0$ as the initial clean image
\For{$k=0,\ldots,K-1$}
    \State Compute attribution $A^k$ from the current clean image $X_k$
    \State Build original mask $M_{\mathrm{orig}}^k$ using Eq.~\eqref{eq:orig_mask}
    \State Update union hard mask $M_{\mathrm{hard}}^k\gets M_{\mathrm{hard}}^{k-1}\cup M_{\mathrm{orig}}^k$
    \State Build soft mask $M_{\mathrm{soft}}^k$ using Eq.~\eqref{eq:soft_mask}
    \State Compute $\mathcal{L}_{\mathrm{joint}}^k$ on $X_k$ and obtain $\nabla_{Z_t^k}\mathcal{L}_{\mathrm{joint}}^k$
    \State Compute $\eta_k$ using Eq.~\eqref{eq:adaptive_step}
    \State Update $Z_t^k$ under $M_{\mathrm{hard}}^k$ to obtain $Z_t^{k+1}$ using Eq.~\eqref{eq:masked_update}
    \State Predict $\widehat{X}_{k+1}\gets G_\theta(Z_t^{k+1},t)$
    \State Blend $\widehat{X}_{k+1}$ with $X_0$ using $M_{\mathrm{soft}}^k$ to obtain $X_{k+1}$ using Eq.~\eqref{eq:blending}
    \If{$p_{\mathrm{target}}(X_{k+1})\ge P_{\mathrm{flip}}$}
        \State \Return $X_{\mathrm{cf}}\gets X_{k+1}$
    \EndIf
\EndFor
\State \Return $X_{\mathrm{cf}}\gets X_K$
\end{algorithmic}
\end{algorithm}

\subsection{Clean-Image Update and Early Stopping}

After the noisy-state update, the PMF predictor $G_\theta$ maps the updated noisy state to a clean prediction:
\begin{equation}
\hat{X}_{k+1} = G_\theta(Z_t^{k+1}, t).
\label{eq:pmf_prediction}
\end{equation}
FiRe then blends this prediction with the original image $X_0$ using the soft mask:
\begin{equation}
X_{k+1}
=
M_{\mathrm{soft}}^k \odot \hat{X}_{k+1}
+
(1-M_{\mathrm{soft}}^k)\odot X_0.
\label{eq:blending}
\end{equation}
Inside the soft-mask region, $X_{k+1}$ takes content from the PMF prediction $\widehat{X}_{k+1}$; outside the mask, it keeps the original image $X_0$. The target classifier evaluates this classifier-facing clean image. 
If $p_{\mathrm{target}}(X_{k+1}) \geq P_{\mathrm{flip}}$, FiRe returns $X_{\mathrm{cf}}=X_{k+1}$; otherwise, refinement continues until the budget $K$ is reached. 

Algorithm~\ref{alg:fire} summarizes the full fixed-noise refinement procedure.


\section{Experiments}

\subsection{Experimental Setup}
\label{sec:experimental_setup}

\paragraph{Datasets and tasks.}
We evaluate FiRe on three datasets and five binary counterfactual explanation tasks, covering both natural and medical images. 
For face counterfactuals, we use the aligned $128\times128$ CelebA dataset~\cite{liu2015deep} and the $256\times256$ CelebA-HQ dataset~\cite{karras2017progressive}, and evaluate two attributes on each dataset: \textit{Smile} and \textit{Age}. 
For medical counterfactuals, we use CheXpert~\cite{irvin2019chexpert} and evaluate the \textit{Pacemaker} shortcut-removal task following FastDiME~\cite{weng2024fast}, extended to $512\times512$ resolution.

\paragraph{Evaluation metrics.}
For CelebA and CelebA-HQ, we follow the ACE evaluation protocol~\cite{jeanneret2023adversarial} and report metrics for realism, identity preservation, sparsity, semantic plausibility, decision consistency, and validity. 
FID~\cite{heusel2017gans} and sFID~\cite{jeanneret2023adversarial} evaluate realism, with sFID reducing the influence of unchanged regions. 
FVA and FS~\cite{cao2018vggface2} measure identity preservation and face similarity, while MNAC~\cite{rodriguez2021beyond} measures non-target attribute changes. 
CD~\cite{jeanneret2022diffusion} evaluates consistency with training-data attribute correlations, COUT~\cite{khorram2022cycle} measures the classifier-output transition, and FR measures whether the counterfactual is classified as the target class. 
For CheXpert, we follow the FastDiME medical-image protocol~\cite{weng2024fast} and report FID~\cite{heusel2017gans}, L1, MAD, S$^3$~\cite{chen2021exploring}, and FR.

\begin{figure}[t]
    \centering
    \includegraphics[width=0.9\linewidth]{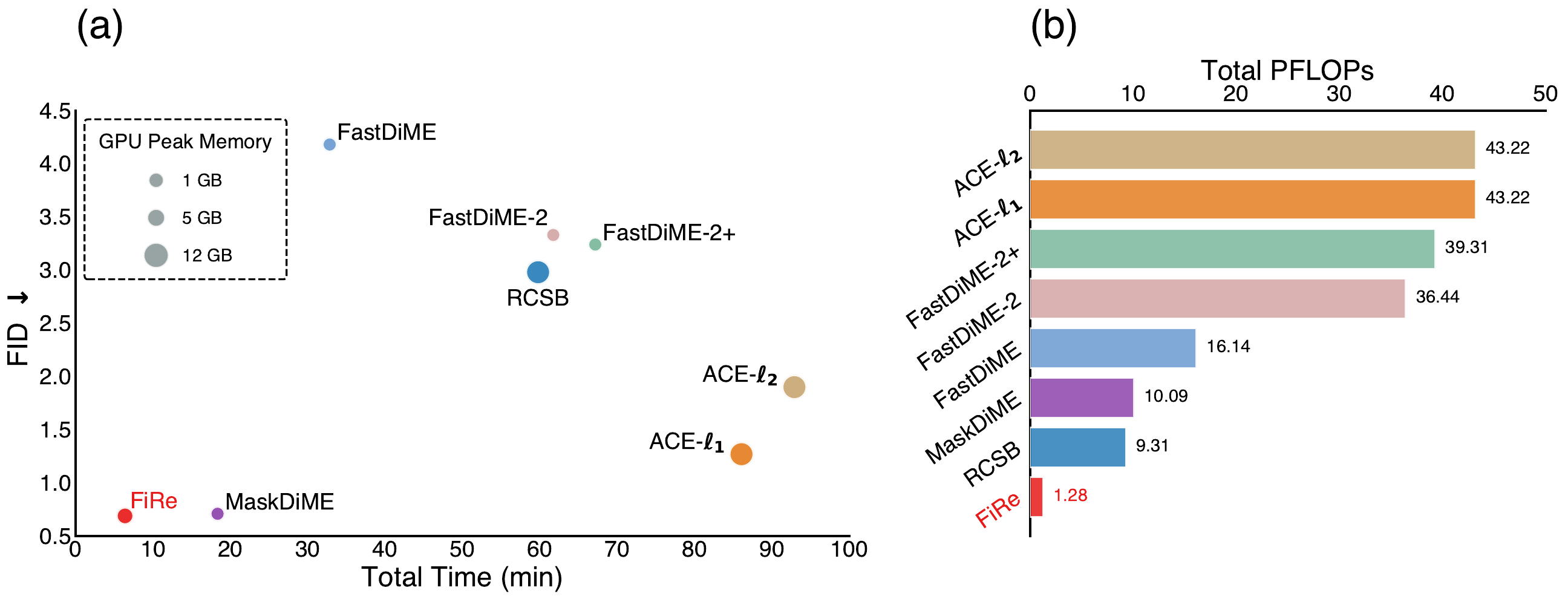}

\caption{\textbf{Computational cost and generation quality comparison} for generating 1,000 counterfactual explanations on the CelebA Smile task (Batch size = 5). 
(a) Practical inference time vs. FID, with marker size indicating peak GPU memory. 
(b) Theoretical compute comparison (Total Peta Floating-Point Operations). 
FiRe achieves the lowest FID while requiring only 6.37 min and 1.28 PFLOPs, yielding 3$\times$ faster inference and 8$\times$ fewer FLOPs than the strongest recent baseline, MaskDiME.}

    \label{fig:dual}
\end{figure}

\paragraph{Comparison protocol.}
For CelebA and CelebA-HQ, we use the same target classifiers and evaluation setup as ACE~\cite{jeanneret2023adversarial}, and train only the unconditional PMF model required by FiRe. 
All methods are evaluated with the same data split, target tasks, classifier, and metrics. 
For CheXpert, directly comparable $512\times512$ results and pretrained classifier weights are not available, so we rerun all baselines using the same $512\times512$ data split, the same DenseNet-121 target classifier~\cite{huang2017densely}, and the same evaluation protocol as FastDiME~\cite{weng2024fast}. 
The trained classifier is fixed and used for all counterfactual methods. 
For the efficiency comparison, we rerun all methods on the CelebA \textit{Smile} task using the same hardware, batch size, classifier, and input resolution. 
Runtime includes counterfactual generation, classifier guidance, and mask construction when applicable, but excludes data loading, disk I/O, and image saving. 
We report the average runtime over the same set of test images for all methods.

\paragraph{Implementation details.}
Unless otherwise specified, FiRe uses the same fixed-noise refinement setting across tasks. 
We set the fixed noise level to $t=0.4$, use one PMF solver step, and allow at most $K=15$ refinement updates. 
The early stopping threshold is $P_{\mathrm{flip}}=0.85$, and the base guidance scale is $\eta=0.02$. 
We use mean-absolute gradient normalization and SmoothGrad with $N=20$ parallel perturbed copies to stabilize attribution. 
The loss weights are $\lambda_{\mathrm{cls}}=1.0$, $\lambda_{\mathrm{perc}}=0.20$, and $\lambda_{\mathrm{tv}}=0.01$. 
Masks are built from target-only gradient attribution with dilation radius $r=2$ and feathering width $w=3$. 
The top saliency ratio $\rho$ is task-specific: $5\%$ for CelebA \textit{Smile}, $12\%$ for CelebA \textit{Age}, and $20\%$ for CheXpert \textit{Pacemaker}. FiRe requires training one PMF model per dataset, whereas several DDPM-based baselines reuse available pretrained weights. PMF training takes 20.4, 12.0, and 8.5 wall-clock hours on CelebA, CelebA-HQ, and CheXpert, respectively, using four AMD MI250X GPUs with 64 GB memory each.

\begin{table}[t]
\centering
\small
\setlength{\tabcolsep}{3pt}
\renewcommand{\arraystretch}{0.8}
\caption{Quantitative comparison on \textbf{CelebA} across the \textit{Smile} and \textit{Age} tasks. Best and second-best results are highlighted in \textbf{bold} and \underline{underline}, respectively.}
\label{tab:celeba}
\begin{tabular}{lcccccccc}
\toprule

\multicolumn{9}{c}{\textbf{CelebA}} \\
\midrule

Method & FID$\downarrow$ & sFID$\downarrow$ & FVA$\uparrow$ & FS$\uparrow$ & MNAC$\downarrow$ & CD$\downarrow$ & COUT$\uparrow$ & FR(\%)$\uparrow$ \\
\midrule

\multicolumn{9}{c}{\textit{Smile}} \\
\midrule
DiVE~\cite{rodriguez2021beyond} & 29.4 & -- & 97.3 & -- & -- & -- & -- & -- \\
DiVE$^{100}$~\cite{rodriguez2021beyond} & 36.8 & -- & 73.4 & -- & 4.63 & 2.34 & -- & -- \\
STEEX~\cite{jacob2022steex} & 10.2 & -- & 96.9 & -- & 4.11 & -- & -- & -- \\
DiME~\cite{jeanneret2022diffusion} & 3.17 & 4.89 & 98.3 & 0.73 & 3.72 & 2.30 & 0.53 & 97.2 \\
ACE~\cite{jeanneret2023adversarial} $\ell_1$ & 1.27 & 3.97 & \underline{99.9} & 0.87 & 2.94 & \underline{1.73} & \underline{0.78} & 97.6 \\
ACE $\ell_2$~\cite{jeanneret2023adversarial} & 1.90 & 4.56 & \underline{99.9} & 0.87 & 2.77 & \textbf{1.56} & 0.62 & 84.3 \\
FastDiME~\cite{weng2024fast} & 4.18 & 6.13 & 99.8 & 0.76 & 3.12 & 1.91 & 0.45 & 99.0 \\
FastDiME-2~\cite{weng2024fast} & 3.33 & 5.49 & \underline{99.9} & 0.77 & 3.06 & 1.89 & 0.44 & 99.4 \\
FastDiME-2+~\cite{weng2024fast} & 3.24 & 5.23 & \underline{99.9} & 0.79 & 2.91 & 2.02 & 0.41 & 98.9 \\
RCSB~\cite{sobieski2025rethinking} & 2.98 & 4.79 & \textbf{100.0} & \underline{0.91} & \underline{2.24} & 2.78 & \textbf{0.87} & \underline{99.8} \\
MaskDiME~\cite{guo2026maskdime} & \underline{0.71} & \underline{3.29} & \textbf{100.0} & \underline{0.91} & 2.78 & 2.41 & \textbf{0.87} & \textbf{100.0} \\
\rowcolor{gray!10}
FiRe (Ours) & \textbf{0.69} & \textbf{3.14} & \textbf{100.0} & \textbf{0.93} & \textbf{1.71} & 2.59 & 0.57 & \textbf{100.0} \\

\midrule

\multicolumn{9}{c}{\textit{Age}} \\
\midrule
DiVE~\cite{rodriguez2021beyond} & 33.8 & -- & 98.2 & -- & 4.58 & -- & -- & -- \\
DiVE$^{100}$~\cite{rodriguez2021beyond} & 39.9 & -- & 52.2 & -- & 4.27 & -- & -- & -- \\
STEEX~\cite{jacob2022steex} & 11.8 & -- & 97.5 & -- & 3.44 & -- & -- & -- \\
DiME~\cite{jeanneret2022diffusion} & 4.15 & 5.89 & 95.3 & 0.67 & 3.13 & 3.27 & 0.44 & 99.0 \\
ACE~\cite{jeanneret2023adversarial} $\ell_1$ & 1.45 & 4.12 & 99.6 & 0.78 & 3.20 & \underline{2.94} & 0.72 & 96.2 \\
ACE $\ell_2$~\cite{jeanneret2023adversarial} & 2.08 & 4.62 & 99.6 & 0.80 & 2.94 & \textbf{2.82} & 0.56 & 77.5 \\
FastDiME~\cite{weng2024fast} & 4.82 & 6.76 & 99.2 & 0.74 & 2.65 & 3.80 & 0.36 & 98.6 \\
FastDiME-2~\cite{weng2024fast} & 4.04 & 6.01 & 99.6 & 0.75 & 2.63 & 3.80 & 0.37 & 99.3 \\
FastDiME-2+~\cite{weng2024fast} & 3.60 & 5.59 & 99.7 & 0.77 & 2.44 & 3.76 & 0.32 & 98.7 \\
RCSB~\cite{sobieski2025rethinking} & 2.94 & 4.94 & \underline{99.9} & \textbf{0.88} & \underline{2.14} & 3.63 & \textbf{0.81} & 99.3 \\
MaskDiME~\cite{guo2026maskdime} & \underline{0.77} & \underline{3.33} & \textbf{100.0} & 0.83 & 2.22 & 3.21 & \underline{0.79} & \textbf{100.0} \\
\rowcolor{gray!10}
FiRe (Ours) & \textbf{0.73} & \textbf{3.29} & \textbf{100.0} & \underline{0.85} & \textbf{1.81} & 3.43 & 0.54 & \textbf{100.0} \\
\bottomrule
\end{tabular}
\end{table}

\subsection{Efficiency Comparison}
\label{sec:efficiency_comparison}

Figure~\ref{fig:dual} compares practical runtime, theoretical compute, and generation quality on the CelebA \textit{Smile} task. All methods are rerun under the same batch size and hardware setting, and runtime measures only counterfactual generation from input images to generated counterfactuals, excluding data loading, disk I/O, and image saving. As shown in Fig.~\ref{fig:dual}(a), FiRe lies in the lower-left region, achieving the best quality-efficiency trade-off among the compared methods. It generates 1,000 counterfactual explanations in 6.37 minutes while also obtaining the lowest FID. Fig.~\ref{fig:dual}(b) shows the same trend in theoretical compute. For generating 1,000 counterfactual explanations, FiRe requires only 1.28 PFLOPs for the full classifier-guided counterfactual generation process, including refinement updates, classifier guidance, and mask construction when applicable. This is about 8$\times$ fewer FLOPs than the strongest recent DDPM-based baseline, MaskDiME.

This efficiency comes from the fixed-noise refinement design. DDPM-based baselines~\cite{jeanneret2022diffusion,jeanneret2023adversarial,weng2024fast,guo2026maskdime} perform classifier-guided editing along reverse diffusion trajectories or require repeated clean-image estimation for guidance. In contrast, FiRe refines the noisy state at a fixed noise level and obtains a clean image from the PMF predictor at each refinement step. Its cost therefore scales with a small number of refinement updates rather than the length of a diffusion trajectory. Although FiRe is not the most memory-light method, its peak GPU memory remains moderate, and the reduction in runtime and FLOPs is substantial.

\begin{table}[t]
\centering
\small
\setlength{\tabcolsep}{3pt}
\renewcommand{\arraystretch}{0.8}
\caption{Quantitative comparison on \textbf{CelebA-HQ} across the \textit{Smile} and \textit{Age} tasks. Best and second-best results are highlighted in \textbf{bold} and \underline{underline}, respectively.}
\label{tab:celeba_hq}
\begin{tabular}{lcccccccc}
\toprule

\multicolumn{9}{c}{\textbf{CelebA-HQ}} \\
\midrule

Method & FID$\downarrow$ & sFID$\downarrow$ & FVA$\uparrow$ & FS$\uparrow$ & MNAC$\downarrow$ & CD$\downarrow$ & COUT$\uparrow$ & FR(\%)$\uparrow$ \\
\midrule

\multicolumn{9}{c}{\textit{Smile}} \\
\midrule
DiVE~\cite{rodriguez2021beyond} & 107.0 & -- & 35.7 & -- & 7.41 & -- & -- & -- \\
STEEX~\cite{jacob2022steex} & 21.9 & -- & 97.6 & -- & 5.27 & -- & -- & -- \\
DiME~\cite{jeanneret2022diffusion} & 18.10 & 27.7 & 96.7 & 0.67 & 2.63 & \underline{1.82} & 0.65 & 97.0 \\
ACE~\cite{jeanneret2023adversarial} $\ell_1$ & 3.21 & 20.2 & \textbf{100.0} & 0.89 & 1.56 & 2.61 & 0.55 & 95.0 \\
ACE $\ell_2$~\cite{jeanneret2023adversarial} & 6.93 & 22.0 & \textbf{100.0} & 0.84 & 1.87 & 2.21 & 0.60 & 95.0 \\
LDCE-txt\cite{farid2023latent} & 13.6 & 25.8 & \underline{99.1} & 0.76 & 2.44 & \textbf{1.68} & 0.34 & -- \\
TiME~\cite{jeanneret2024text} & 10.98 & 23.8 & 96.6 & 0.79 & 2.97 & 2.32 & 0.63 & 97.1 \\
RCSB~\cite{sobieski2025rethinking} & 3.04 & 20.0 & \textbf{100.0} & \underline{0.93} & \textbf{1.22} & 3.22 & \textbf{0.83} & 98.9 \\
MaskDiME~\cite{guo2026maskdime} & \underline{2.51} & \underline{18.1} & \textbf{100.0} & \textbf{0.94} & \underline{1.41} & 2.67 & \underline{0.69} & \underline{99.4} \\
\rowcolor{gray!10}
FiRe (Ours) & \textbf{1.98} & \textbf{17.9} & \textbf{100.0} & \underline{0.93} & \textbf{1.22} & 2.58 & 0.62 & \textbf{99.9} \\

\midrule

\multicolumn{9}{c}{\textit{Age}} \\
\midrule
DiVE~\cite{rodriguez2021beyond} & 107.5 & -- & 32.3 & -- & 6.76 & -- & -- & -- \\
STEEX~\cite{jacob2022steex} & 26.8 & -- & -- & -- & 5.63 & -- & -- & -- \\
DiME~\cite{jeanneret2022diffusion} & 18.7 & 27.8 & 95.0 & 0.66 & 2.10 & 4.29 & 0.56 & 97.0 \\
ACE~\cite{jeanneret2023adversarial} $\ell_1$ & 5.31 & 21.7 & \underline{99.6} & 0.81 & 1.53 & 5.40 & 0.40 & 95.0 \\
ACE $\ell_2$~\cite{jeanneret2023adversarial} & 16.4 & 28.2 & \underline{99.6} & 0.77 & 1.92 & \underline{4.21} & 0.53 & 95.0 \\
LDCE-txt\cite{farid2023latent} & 14.2 & 25.6 & 98.0 & 0.73 & 2.12 & \textbf{4.02} & 0.33 & -- \\
TiME~\cite{jeanneret2024text} & 20.9 & 32.9 & 79.3 & 0.63 & 4.19 & 4.29 & 0.31 & 89.9 \\
RCSB~\cite{sobieski2025rethinking} & 4.92 & 27.3 & \textbf{100.0} & \textbf{0.96} & \underline{1.47} & 5.16 & \textbf{0.80} & 99.4 \\
MaskDiME~\cite{guo2026maskdime} & \underline{4.43} & \underline{19.4} & \textbf{100.0} & 0.88 & 1.82 & 4.67 & \underline{0.63} & \underline{99.6} \\
\rowcolor{gray!10}
FiRe (Ours) & \textbf{3.02} & \textbf{18.6} & \textbf{100.0} & \underline{0.90} & \textbf{1.46} & 4.72 & 0.55 & \textbf{99.9} \\

\bottomrule
\end{tabular}
\end{table}

\subsection{Quantitative Comparison}
\label{sec:quantitative_comparison}

\begin{wraptable}{r}{0.55\textwidth}
\centering
\vspace{-8pt}
\caption{Quantitative comparison on \textbf{CheXpert} for the \textit{Pacemaker} task. Best and second-best results are highlighted in \textbf{bold} and \underline{underline}, respectively.}
\label{tab:chexpert}

\footnotesize
\setlength{\tabcolsep}{3pt}
\renewcommand{\arraystretch}{1.05}

\begin{tabular}{lccccc}
\toprule
\multicolumn{6}{c}{\textbf{CheXpert}} \\
\midrule
Method & FID$\downarrow$ & L1$\downarrow$ & MAD$\uparrow$ & S$^3$$\uparrow$ & FR(\%)$\uparrow$ \\
\midrule
\multicolumn{6}{c}{\textit{Pacemaker}} \\
\midrule
DiME~\cite{jeanneret2022diffusion}        & 51.7 & 0.088 & 0.956 & 0.902 & 97.7 \\
FastDiME~\cite{weng2024fast}    & 29.2 & 0.058 & \underline{0.959} & 0.927 & \textbf{100.0} \\
FastDiME-2~\cite{weng2024fast}  & 30.4 & 0.069 & 0.953 & 0.881 & \underline{99.8} \\
FastDiME-2+~\cite{weng2024fast} & 38.1 & \underline{0.052} & 0.954 & 0.928 & \textbf{100.0} \\
MaskDiME~\cite{guo2026maskdime}    & \underline{22.0} & 0.053 & 0.954 & \underline{0.935} & \textbf{100.0} \\
\rowcolor{gray!10}
FiRe (Ours) & \textbf{17.4} & \textbf{0.049} & \textbf{0.960} & \textbf{0.951} & \textbf{100.0} \\
\bottomrule
\end{tabular}

\vspace{-8pt}
\end{wraptable}

Tables~\ref{tab:celeba}, \ref{tab:celeba_hq}, and~\ref{tab:chexpert} compare FiRe with prior VCE methods on face counterfactuals and a medical shortcut-removal task. Overall, FiRe improves realism and edit sparsity while maintaining high counterfactual validity, showing that the efficiency gain from fixed-noise refinement does not come at the cost of counterfactual quality or explanation reliability.

On CelebA and CelebA-HQ, Tables~\ref{tab:celeba} and~\ref{tab:celeba_hq} evaluate FiRe on four face counterfactual tasks at $128\times128$ and $256\times256$ resolutions. FiRe achieves the best FID and sFID on all four tasks, indicating stronger realism under the reported distributional metrics. It also obtains the best or tied-best MNAC in most cases while maintaining near-perfect FR, showing that FiRe reaches the target decision with fewer unintended changes to non-target attributes. This supports localized refinement in avoiding unnecessary global edits. On CelebA-HQ, FiRe keeps FVA and FS competitive with the strongest baselines, suggesting that improved FID and sFID do not come with a clear drop in identity-related similarity metrics. FiRe does not achieve the best CD or COUT on all face tasks. This behavior is consistent with its design goal. Once the classifier reaches the target decision, FiRe keeps the earliest sufficient counterfactual rather than continuing refinement to maximize classifier output shift or induce stronger correlated attribute changes. Lower COUT in some cases should therefore be read together with FiRe's high FR, low FID/sFID, and low MNAC: the method favors sufficient and localized counterfactual edits over stronger but potentially less sparse decision shifts. A direct ablation of this trade-off is provided in Table~\ref{tab:ablation}.

On CheXpert, Table~\ref{tab:chexpert} evaluates FiRe on the \textit{Pacemaker} shortcut-removal task at $512\times512$ resolution. FiRe achieves the best FID, L1, MAD, and S$^3$, and reaches $100.0\%$ FR. These results indicate that FiRe can remove the shortcut cue while preserving pixel-level closeness and representation-level similarity. This suggests that the edit remains localized to the classifier-relevant cue. This experiment does not claim broad medical generalization, but shows that FiRe transfers well to a higher-resolution medical shortcut-removal setting with a data distribution different from face images.

\subsection{Qualitative Comparison}
\label{sec:qualitative_comparison}

\begin{figure}[t]
    \centering
    \includegraphics[width=0.9\linewidth]{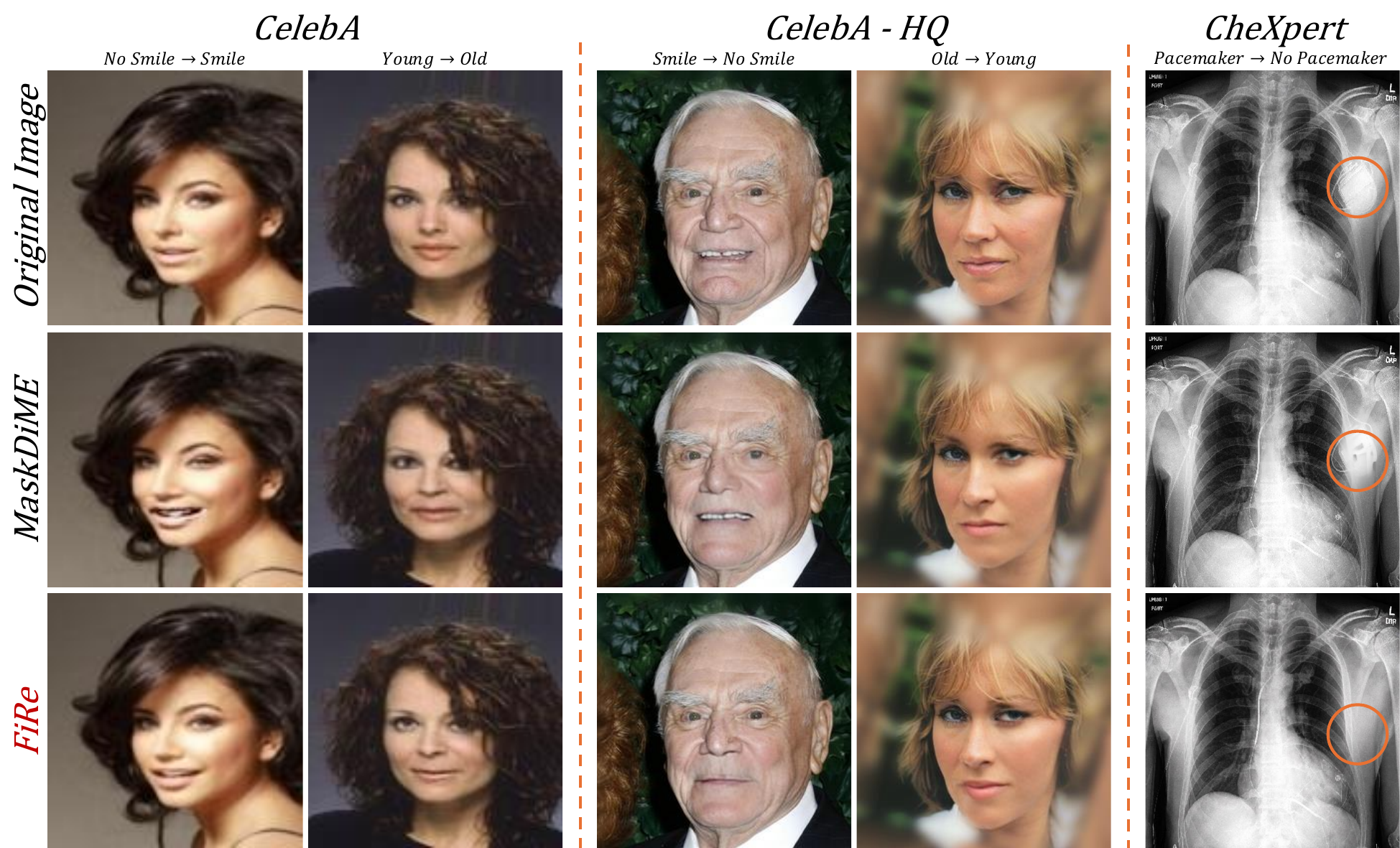}

\caption{\textbf{Qualitative comparison.} FiRe makes the target change while preserving non-target facial regions and surrounding anatomical structure better than MaskDiME.}

    \label{fig:qualitative}
\end{figure}

Figure~\ref{fig:qualitative} shows a qualitative comparison between FiRe and MaskDiME~\cite{guo2026maskdime}, the strongest recent DDPM-based baseline, across CelebA, CelebA-HQ, and CheXpert.
On CelebA, both methods change the target attributes for \textit{No Smile} $\rightarrow$ \textit{Smile} and \textit{Young} $\rightarrow$ \textit{Old}. FiRe produces these changes with less visible alteration to non-target facial regions, such as hair, background, and overall face structure. On CelebA-HQ, FiRe similarly keeps the global face layout and fine facial details closer to the original image for \textit{Smile} $\rightarrow$ \textit{No Smile} and \textit{Old} $\rightarrow$ \textit{Young}, where unnecessary edits are easier to observe. On CheXpert, FiRe removes the highlighted pacemaker cue while leaving fewer residual artifacts around the edited region and preserving nearby anatomical structure. These examples align with the quantitative results in Tables~\ref{tab:celeba}, \ref{tab:celeba_hq}, and~\ref{tab:chexpert}, where FiRe improves realism and sparsity metrics while maintaining high counterfactual validity, supporting its goal of sufficient but localized editing.

\subsection{Ablation Study}
\label{sec:ablation_study}

\begin{table}[t]
\centering
\scriptsize
\setlength{\tabcolsep}{3.5pt}
\caption{\textbf{Main ablation on CelebA-HQ Smile.} We compare three PMF-based formulations: one-step guidance, reverse refinement with changing noise levels, and fixed-noise refinement. We then compare fixed-noise refinement using a MaskDiME-style mask, the FiRe mask, adaptive guidance without early stopping, and the full FiRe model with early stopping.}
\label{tab:ablation}
\resizebox{\linewidth}{!}{
\begin{tabular}{lcccccccc}
\toprule
Method & FID$\downarrow$ & sFID$\downarrow$ & FVA$\uparrow$ & FS$\uparrow$ & MNAC$\downarrow$ & CD$\downarrow$ & COUT$\uparrow$ & FR(\%)$\uparrow$ \\
\midrule
PMF-One Step & 27.36 & 38.5 & 97.4 & 0.84 & 3.40 & 3.11 & 0.67 & 41.6 \\
PMF-Reverse & 31.49 & 42.2 & 97.4 & 0.85 & 2.98 & 2.94 & 0.81 & 90.3 \\
PMF-Fixed & 19.13 & 30.2 & 100.0 & 0.88 & 2.95 & 2.41 & 0.88 & 95.2 \\
\midrule
PMF-Fixed w/ MaskDiME Mask & 9.41 & 25.8 & 99.8 & 0.88 & 1.75 & 2.63 & 0.71 & 93.7 \\
PMF-Fixed w/ FiRe Mask & 3.97 & 24.3 & 100.0& 0.91 & 1.44 & 2.46 & 0.68 & 99.9 \\
FiRe w/o Early Stop & 3.47 & 22.9 & 100.0 & 0.91 & 1.42 & 2.51 & 0.66 & 99.9 \\
\rowcolor{gray!10}
FiRe  & 1.98 & 17.9 & 100.0 & 0.93 & 1.22 & 2.58 & 0.62 & 99.9 \\
\bottomrule
\end{tabular}
}
\end{table}

\begin{table}[t]
\centering

\begin{minipage}{0.48\linewidth}
\centering
\caption{Sensitivity analysis of the refinement budget $K$ on CelebA-HQ \textit{Smile}.}
\label{tab:k_sensitivity}
\footnotesize
\setlength{\tabcolsep}{3pt}
\renewcommand{\arraystretch}{1.05}
\begin{tabular}{lccccc}
\toprule
$K$ & FID$\downarrow$ & sFID$\downarrow$ & FS$\uparrow$ & MNAC$\downarrow$ & FR(\%)$\uparrow$ \\
\midrule
5  & 3.60 & 21.4 & 0.91 & 1.41 & 92.4 \\
10 & 2.44 & 19.5 & 0.93 & 1.28 & 99.1 \\
\rowcolor{gray!10}
15 & 1.98 & 17.9 & 0.93 & 1.22 & 99.9 \\
20 & 2.25 & 18.6 & 0.90 & 1.25 & 99.9 \\
\bottomrule
\end{tabular}
\end{minipage}
\hfill
\begin{minipage}{0.48\linewidth}
\centering
\caption{Sensitivity analysis of the fixed noise level $t$ on CelebA-HQ \textit{Smile}.}
\label{tab:t_sensitivity}
\footnotesize
\setlength{\tabcolsep}{3pt}
\renewcommand{\arraystretch}{1.05}
\begin{tabular}{lccccc}
\toprule
$t$ & FID$\downarrow$ & sFID$\downarrow$ & FS$\uparrow$ & MNAC$\downarrow$ & FR(\%)$\uparrow$ \\
\midrule
0.2 & 5.37 & 23.8 & 0.87 & 3.16 & 87.3 \\
\rowcolor{gray!10}
0.4 & 1.98 & 17.9 & 0.93 & 1.22 & 99.9 \\
0.6 & 8.14 & 27.2 & 0.91 & 1.78 & 99.9 \\
0.8 & 12.6 & 35.1 & 0.87 & 2.20 & 99.9 \\
\bottomrule
\end{tabular}
\end{minipage}

\end{table}

\begin{figure}[t]
    \centering
    \includegraphics[width=0.9\linewidth]{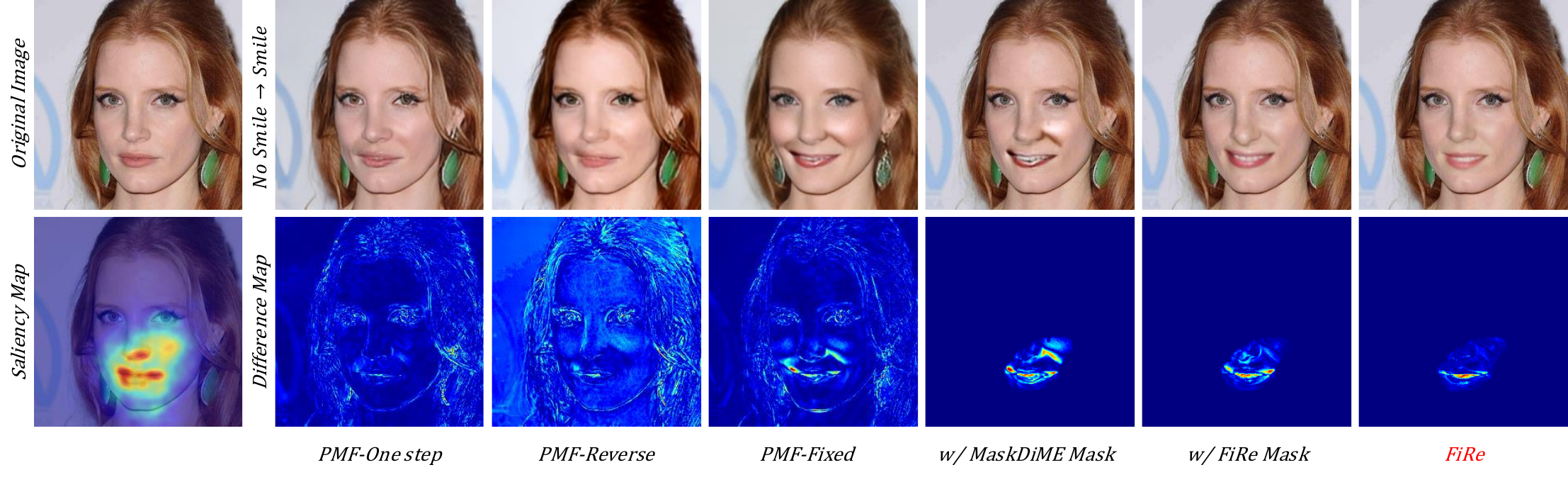}

\caption{
\textbf{Qualitative ablation on the CelebA-HQ \textit{Smile} task.}
PMF-One Step produces an insufficient target edit.
PMF-Reverse strengthens the smile edit but introduces broad modifications across the face and background, while PMF-Fixed reduces part of this drift but remains spatially diffuse.
A MaskDiME mask improves locality but still leaves scattered changes and artifacts.
The FiRe mask concentrates the edit around the mouth, but spatial control alone cannot prevent overly strong updates within the selected region. 
Adaptive guidance and early stopping further regulate the refinement process, reducing unnecessary modifications while preserving a valid smile counterfactual.
}

    \label{fig:ablation}
\end{figure}

Table~\ref{tab:ablation} and Figure~\ref{fig:ablation} provide a progressive ablation of FiRe on the CelebA-HQ \textit{Smile} task. 
Instead of removing individual components from the full model, we start from a basic PMF-guided counterfactual generator and progressively build toward the complete fixed-noise refinement framework. 
This design examines whether PMF clean-image prediction~\cite{lu2026one} is sufficient to support classifier-guided counterfactual explanation, whether fixed-noise refinement is more suitable than a DDPM-based reverse denoising trajectory~\cite{ho2020denoising,jeanneret2022diffusion}, and whether FiRe-specific masking, adaptive guidance, and early stopping are necessary for localized and sufficient counterfactual edits.

PMF-One Step uses PMF only as a clean-image interface and applies classifier guidance once. 
Although PMF can directly provide a classifier-facing clean image, one-step guidance is insufficient to reliably cross the classifier decision boundary, leading to a low flip rate and weak smile edits in the qualitative results. 
PMF-Reverse performs 15-step refinement along a DDPM-based reverse denoising trajectory and substantially improves the flip rate. 
However, the difference map in Figure~\ref{fig:ablation} shows that its modifications are not limited to the mouth region, but also spread strongly to hair and background regions. 
In contrast, PMF-Fixed keeps all 15 refinement steps at the same noise level. 
Compared with PMF-Reverse, it achieves better quantitative results and produces edits that are more concentrated around the mouth, indicating that fixed-noise refinement is a more suitable formulation for PMF-guided counterfactual explanation. 
Nevertheless, PMF-Fixed still lacks explicit spatial control, so its modifications remain insufficiently clean and localized.

We then analyze the spatial and refinement controls in FiRe. 
Adding a MaskDiME-style mask~\cite{guo2026maskdime} constrains the edited region to some extent and improves locality, but the qualitative results still show scattered artifacts. 
This suggests that trajectory-style masking is not fully compatible with fixed-noise refinement. 
FiRe therefore adopts a dual-mask design: the hard-mask memory constrains noisy-state updates and preserves edited regions across refinement steps, while the soft visible mask controls the clean-image update and smooths visible edit boundaries. 
This design improves realism, sparsity, and validity, and further concentrates the edit around the mouth region. 
Finally, we isolate early stopping by comparing FiRe with a variant that retains adaptive guidance but disables early stopping. Without early stopping, COUT increases from 0.62 to 0.66 while FR remains 99.9\%, but FID degrades from 1.98 to 3.47, sFID from 17.9 to 22.9, and MNAC from 1.22 to 1.42. This result indicates that early stopping favors the earliest sufficient counterfactual, improving realism and sparsity at the cost of a modest reduction in classifier-output transition.

Tables~\ref{tab:k_sensitivity} and~\ref{tab:t_sensitivity} analyze sensitivity to the refinement budget $K$ and fixed noise level $t$. Increasing $K$ from 5 to 15 improves both realism and validity, while increasing it to 20 gives no additional benefit and slightly degrades proximity. In practice, over 99\% of successful counterfactuals stop within the first 10 refinement steps, and only a very small fraction requires the remaining steps. This suggests that $K=15$ is a safe maximum budget, while early stopping avoids unnecessary refinement after the target decision is reached. For the noise level, a small value such as $t=0.2$ preserves the input but limits editability, resulting in a lower flip rate. Larger values such as $t=0.6$ and $t=0.8$ keep high validity but degrade realism and sparsity. We therefore use $K=15$ and $t=0.4$ as the default setting, which provides the best overall trade-off.

\section{Conclusion}
\label{sec:conclusion}

We introduced FiRe, a fixed-noise refinement framework for classifier-guided visual counterfactual explanation. FiRe shows that counterfactual editing does not need to be formulated as guidance along changing DDPM reverse timesteps. Instead, refining a noisy state at a single intermediate noise level provides a more stable balance between semantic editability and spatial control. Together with PMF-based clean-image prediction, this formulation enables efficient and reliable classifier-facing guidance without relying on recursive denoising or lower-quality one-step DDPM estimates. FiRe further uses dynamic dual masking, adaptive guidance, and early stopping to keep edits localized and sufficient. Experiments on CelebA, CelebA-HQ, and CheXpert demonstrate that FiRe substantially improves efficiency while preserving competitive or state-of-the-art counterfactual quality. These results show fixed-noise refinement as a promising direction for efficient and controllable diffusion-based visual explanations.

\newpage

\bibliography{egbib}

@article{arrieta2020explainable,
  title     = {Explainable Artificial Intelligence ({XAI}): Concepts, Taxonomies, Opportunities and Challenges Toward Responsible {AI}},
  author    = {Arrieta, Alejandro Barredo and D{\'i}az-Rodr{\'i}guez, Natalia and Del Ser, Javier and Bennetot, Adrien and Tabik, Siham and Barbado, Alberto and Garc{\'i}a, Salvador and Gil-L{\'o}pez, Sergio and Molina, Daniel and Benjamins, Richard and Chatila, Raja and Herrera, Francisco},
  journal   = {Information Fusion},
  volume    = {58},
  pages     = {82--115},
  year      = {2020},
  publisher = {Elsevier}
}

@article{wachter2017counterfactual,
  title     = {Counterfactual Explanations Without Opening the Black Box: Automated Decisions and the {GDPR}},
  author    = {Wachter, Sandra and Mittelstadt, Brent and Russell, Chris},
  journal   = {Harvard Journal of Law \& Technology},
  volume    = {31},
  number    = {2},
  pages     = {841--887},
  year      = {2017}
}

@inproceedings{hendricks2016generating,
  title     = {Generating Visual Explanations},
  author    = {Hendricks, Lisa Anne and Akata, Zeynep and Rohrbach, Marcus and Donahue, Jeff and Schiele, Bernt and Darrell, Trevor},
  booktitle = {European Conference on Computer Vision},
  pages     = {3--19},
  year      = {2016}
}

@inproceedings{goyal2019counterfactual,
  title     = {Counterfactual Visual Explanations},
  author    = {Goyal, Yash and Wu, Ziyan and Ernst, Jan and Batra, Dhruv and Parikh, Devi and Lee, Stefan},
  booktitle = {International Conference on Machine Learning},
  pages     = {2376--2384},
  year      = {2019}
}

@inproceedings{rodriguez2021beyond,
  title     = {Beyond Trivial Counterfactual Explanations with Diverse Valuable Explanations},
  author    = {Rodriguez, Pau and Caccia, Massimo and Lacoste, Alexandre and Zamparo, Lee and Laradji, Issam and Charlin, Laurent and Vazquez, David},
  booktitle = {Proceedings of the IEEE/CVF International Conference on Computer Vision},
  pages     = {1056--1065},
  year      = {2021}
}

@inproceedings{jacob2022steex,
  title     = {{STEEX}: Steering Counterfactual Explanations with Semantics},
  author    = {Jacob, Paul and Zablocki, {\'E}loi and Ben-Younes, Hedi and Chen, Micka{\"e}l and P{\'e}rez, Patrick and Cord, Matthieu},
  booktitle = {European Conference on Computer Vision},
  pages     = {387--403},
  year      = {2022}
}

@article{joshi2018xgems,
  title   = {{xGEMs}: Generating Exemplars to Explain Black-Box Models},
  author  = {Joshi, Shalmali and Koyejo, Oluwasanmi and Kim, Been and Ghosh, Joydeep},
  journal = {arXiv preprint arXiv:1806.08867},
  year    = {2018}
}

@inproceedings{jeanneret2022diffusion,
  title     = {Diffusion Models for Counterfactual Explanations},
  author    = {Jeanneret, Guillaume and Simon, Lo{\"\i}c and Jurie, Fr{\'e}d{\'e}ric},
  booktitle = {Proceedings of the Asian Conference on Computer Vision},
  pages     = {858--876},
  year      = {2022}
}

@article{augustin2022diffusion,
  title   = {Diffusion Visual Counterfactual Explanations},
  author  = {Augustin, Maximilian and Boreiko, Valentyn and Croce, Francesco and Hein, Matthias},
  journal = {Advances in Neural Information Processing Systems},
  volume  = {35},
  pages   = {364--377},
  year    = {2022}
}

@inproceedings{jeanneret2023adversarial,
  title     = {Adversarial Counterfactual Visual Explanations},
  author    = {Jeanneret, Guillaume and Simon, Lo{\"\i}c and Jurie, Fr{\'e}d{\'e}ric},
  booktitle = {Proceedings of the IEEE/CVF Conference on Computer Vision and Pattern Recognition},
  pages     = {16425--16435},
  year      = {2023}
}

@inproceedings{weng2024fast,
  title     = {Fast Diffusion-Based Counterfactuals for Shortcut Removal and Generation},
  author    = {Weng, Nina and Pegios, Paraskevas and Petersen, Eike and Feragen, Aasa and Bigdeli, Siavash},
  booktitle = {European Conference on Computer Vision},
  pages     = {338--357},
  year      = {2024}
}

@article{guo2026maskdime,
  title   = {{MaskDiME}: Adaptive Masked Diffusion for Precise and Efficient Visual Counterfactual Explanations},
  author  = {Guo, Changlu and Christensen, Anders Nymark and Dahl, Anders Bjorholm and Hannemose, Morten Rieger},
  journal = {arXiv preprint arXiv:2602.18792},
  year    = {2026}
}

@inproceedings{sobieski2025rethinking,
  title     = {Rethinking Visual Counterfactual Explanations Through Region Constraint},
  author    = {Sobieski, Bartlomiej and Grzywaczewski, Jakub and Sadlej, Bart{\l}omiej and Tivnan, Matthew and Biecek, Przemyslaw},
  booktitle = {International Conference on Learning Representations},
  year      = {2025}
}

@article{farid2023latent,
  title   = {Latent Diffusion Counterfactual Explanations},
  author  = {Farid, Karim and Schrodi, Simon and Argus, Max and Brox, Thomas},
  journal = {arXiv preprint arXiv:2310.06668},
  year    = {2023}
}

@inproceedings{jeanneret2024text,
  title     = {Text-to-Image Models for Counterfactual Explanations: A Black-Box Approach},
  author    = {Jeanneret, Guillaume and Simon, Lo{\"\i}c and Jurie, Fr{\'e}d{\'e}ric},
  booktitle = {Proceedings of the IEEE/CVF Winter Conference on Applications of Computer Vision},
  pages     = {4757--4767},
  year      = {2024}
}

@article{kumar2025prism,
  title   = {{PRISM}: High-Resolution and Precise Counterfactual Medical Image Generation Using Language-Guided Stable Diffusion},
  author  = {Kumar, Amar and Kriz, Anita and Havaei, Mohammad and Arbel, Tal},
  journal = {arXiv preprint arXiv:2503.00196},
  year    = {2025}
}

@article{fathi2024decodex,
  title   = {{DeCoDEx}: Confounder Detector Guidance for Improved Diffusion-Based Counterfactual Explanations},
  author  = {Fathi, Nima and Kumar, Amar and Nichyporuk, Brennan and Havaei, Mohammad and Arbel, Tal},
  journal = {arXiv preprint arXiv:2405.09288},
  year    = {2024}
}

@article{ho2020denoising,
  title   = {Denoising Diffusion Probabilistic Models},
  author  = {Ho, Jonathan and Jain, Ajay and Abbeel, Pieter},
  journal = {Advances in Neural Information Processing Systems},
  volume  = {33},
  pages   = {6840--6851},
  year    = {2020}
}

@article{dhariwal2021diffusion,
  title   = {Diffusion Models Beat {GANs} on Image Synthesis},
  author  = {Dhariwal, Prafulla and Nichol, Alexander},
  journal = {Advances in Neural Information Processing Systems},
  volume  = {34},
  pages   = {8780--8794},
  year    = {2021}
}

@inproceedings{rombach2022high,
  title     = {High-Resolution Image Synthesis with Latent Diffusion Models},
  author    = {Rombach, Robin and Blattmann, Andreas and Lorenz, Dominik and Esser, Patrick and Ommer, Bj{\"o}rn},
  booktitle = {Proceedings of the IEEE/CVF Conference on Computer Vision and Pattern Recognition},
  pages     = {10684--10695},
  year      = {2022}
}

@incollection{robbins1992empirical,
  title     = {An Empirical {Bayes} Approach to Statistics},
  author    = {Robbins, Herbert E.},
  booktitle = {Breakthroughs in Statistics: Foundations and Basic Theory},
  pages     = {388--394},
  publisher = {Springer},
  year      = {1992}
}

@article{lipman2022flow,
  title   = {Flow Matching for Generative Modeling},
  author  = {Lipman, Yaron and Chen, Ricky T. Q. and Ben-Hamu, Heli and Nickel, Maximilian and Le, Matt},
  journal = {arXiv preprint arXiv:2210.02747},
  year    = {2022}
}

@article{lu2026one,
  title   = {One-Step Latent-Free Image Generation with {Pixel Mean Flows}},
  author  = {Lu, Yiyang and Lu, Susie and Sun, Qiao and Zhao, Hanhong and Jiang, Zhicheng and Wang, Xianbang and Li, Tianhong and Geng, Zhengyang and He, Kaiming},
  journal = {arXiv preprint arXiv:2601.22158},
  year    = {2026}
}

@article{simonyan2013deep,
  title   = {Deep Inside Convolutional Networks: Visualising Image Classification Models and Saliency Maps},
  author  = {Simonyan, Karen and Vedaldi, Andrea and Zisserman, Andrew},
  journal = {arXiv preprint arXiv:1312.6034},
  year    = {2013}
}

@article{smilkov2017smoothgrad,
  title   = {{SmoothGrad}: Removing Noise by Adding Noise},
  author  = {Smilkov, Daniel and Thorat, Nikhil and Kim, Been and Vi{\'e}gas, Fernanda and Wattenberg, Martin},
  journal = {arXiv preprint arXiv:1706.03825},
  year    = {2017}
}

@inproceedings{selvaraju2017grad,
  title     = {{Grad-CAM}: Visual Explanations from Deep Networks via Gradient-Based Localization},
  author    = {Selvaraju, Ramprasaath R. and Cogswell, Michael and Das, Abhishek and Vedantam, Ramakrishna and Parikh, Devi and Batra, Dhruv},
  booktitle = {Proceedings of the IEEE International Conference on Computer Vision},
  pages     = {618--626},
  year      = {2017}
}

@inproceedings{kim2018interpretability,
  title     = {Interpretability Beyond Feature Attribution: Quantitative Testing with Concept Activation Vectors ({TCAV})},
  author    = {Kim, Been and Wattenberg, Martin and Gilmer, Justin and Cai, Carrie and Wexler, James and Vi{\'e}gas, Fernanda and others},
  booktitle = {International Conference on Machine Learning},
  pages     = {2668--2677},
  year      = {2018}
}

@inproceedings{lang2021explaining,
  title     = {Explaining in Style: Training a {GAN} to Explain a Classifier in {StyleSpace}},
  author    = {Lang, Oran and Gandelsman, Yossi and Yarom, Michal and Wald, Yoav and Elidan, Gal and Hassidim, Avinatan and Freeman, William T. and Isola, Phillip and Globerson, Amir and Irani, Michal and others},
  booktitle = {Proceedings of the IEEE/CVF International Conference on Computer Vision},
  pages     = {693--702},
  year      = {2021}
}

@inproceedings{choi2018stargan,
  title     = {{StarGAN}: Unified Generative Adversarial Networks for Multi-Domain Image-to-Image Translation},
  author    = {Choi, Yunjey and Choi, Minje and Kim, Munyoung and Ha, Jung-Woo and Kim, Sunghun and Choo, Jaegul},
  booktitle = {Proceedings of the IEEE Conference on Computer Vision and Pattern Recognition},
  pages     = {8789--8797},
  year      = {2018}
}

@article{he2019attgan,
  title     = {{AttGAN}: Facial Attribute Editing by Only Changing What You Want},
  author    = {He, Zhenliang and Zuo, Wangmeng and Kan, Meina and Shan, Shiguang and Chen, Xilin},
  journal   = {IEEE Transactions on Image Processing},
  volume    = {28},
  number    = {11},
  pages     = {5464--5478},
  year      = {2019},
  publisher = {IEEE}
}

@article{heusel2017gans,
  title   = {{GANs} Trained by a Two Time-Scale Update Rule Converge to a Local {Nash} Equilibrium},
  author  = {Heusel, Martin and Ramsauer, Hubert and Unterthiner, Thomas and Nessler, Bernhard and Hochreiter, Sepp},
  journal = {Advances in Neural Information Processing Systems},
  volume  = {30},
  year    = {2017}
}

@inproceedings{zhang2018unreasonable,
  title     = {The Unreasonable Effectiveness of Deep Features as a Perceptual Metric},
  author    = {Zhang, Richard and Isola, Phillip and Efros, Alexei A. and Shechtman, Eli and Wang, Oliver},
  booktitle = {Proceedings of the IEEE Conference on Computer Vision and Pattern Recognition},
  pages     = {586--595},
  year      = {2018}
}

@inproceedings{johnson2016perceptual,
  title     = {Perceptual Losses for Real-Time Style Transfer and Super-Resolution},
  author    = {Johnson, Justin and Alahi, Alexandre and Fei-Fei, Li},
  booktitle = {European Conference on Computer Vision},
  pages     = {694--711},
  year      = {2016}
}

@inproceedings{liu2015deep,
  title     = {Deep Learning Face Attributes in the Wild},
  author    = {Liu, Ziwei and Luo, Ping and Wang, Xiaogang and Tang, Xiaoou},
  booktitle = {Proceedings of the IEEE International Conference on Computer Vision},
  pages     = {3730--3738},
  year      = {2015}
}

@article{karras2017progressive,
  title   = {Progressive Growing of {GANs} for Improved Quality, Stability, and Variation},
  author  = {Karras, Tero and Aila, Timo and Laine, Samuli and Lehtinen, Jaakko},
  journal = {arXiv preprint arXiv:1710.10196},
  year    = {2017}
}

@inproceedings{irvin2019chexpert,
  title     = {{CheXpert}: A Large Chest Radiograph Dataset with Uncertainty Labels and Expert Comparison},
  author    = {Irvin, Jeremy and Rajpurkar, Pranav and Ko, Michael and Yu, Yifan and Ciurea-Ilcus, Silviana and Chute, Chris and Marklund, Henrik and Haghgoo, Behzad and Ball, Robyn and Shpanskaya, Katie and others},
  booktitle = {Proceedings of the AAAI Conference on Artificial Intelligence},
  volume    = {33},
  number    = {01},
  pages     = {590--597},
  year      = {2019}
}

@inproceedings{chen2021exploring,
  title     = {Exploring Simple {Siamese} Representation Learning},
  author    = {Chen, Xinlei and He, Kaiming},
  booktitle = {Proceedings of the IEEE/CVF Conference on Computer Vision and Pattern Recognition},
  pages     = {15750--15758},
  year      = {2021}
}

@inproceedings{huang2017densely,
  title     = {Densely Connected Convolutional Networks},
  author    = {Huang, Gao and Liu, Zhuang and Van Der Maaten, Laurens and Weinberger, Kilian Q.},
  booktitle = {Proceedings of the IEEE Conference on Computer Vision and Pattern Recognition},
  pages     = {4700--4708},
  year      = {2017}
}

@article{rudin1992nonlinear,
  title   = {Nonlinear Total Variation Based Noise Removal Algorithms},
  author  = {Rudin, Leonid I. and Osher, Stanley and Fatemi, Emad},
  journal = {Physica D: Nonlinear Phenomena},
  volume  = {60},
  number  = {1--4},
  pages   = {259--268},
  year    = {1992}
}

@inproceedings{cao2018vggface2,
  title     = {{VGGFace2}: A Dataset for Recognising Faces Across Pose and Age},
  author    = {Cao, Qiong and Shen, Li and Xie, Weidi and Parkhi, Omkar M. and Zisserman, Andrew},
  booktitle = {Proceedings of the IEEE International Conference on Automatic Face and Gesture Recognition},
  pages     = {67--74},
  year      = {2018}
}

@inproceedings{khorram2022cycle,
  title     = {Cycle-Consistent Counterfactuals by Latent Transformations},
  author    = {Khorram, Saeed and Fuxin, Li},
  booktitle = {Proceedings of the IEEE/CVF Conference on Computer Vision and Pattern Recognition},
  pages     = {10203--10212},
  year      = {2022}
}

@inproceedings{boreiko2022sparse,
  title     = {Sparse Visual Counterfactual Explanations in Image Space},
  author    = {Boreiko, Valentyn and Augustin, Maximilian and Croce, Francesco and Berens, Philipp and Hein, Matthias},
  booktitle = {German Conference on Pattern Recognition},
  pages     = {133--148},
  year      = {2022}
}

@inproceedings{liu2023flow,
  title     = {Flow Straight and Fast: Learning to Generate and Transfer Data with Rectified Flow},
  author    = {Liu, Xingchao and Gong, Chengyue and Liu, Qiang},
  booktitle = {International Conference on Learning Representations},
  year      = {2023}
}

@article{liu2022rectified,
  title   = {Rectified Flow: A Marginal Preserving Approach to Optimal Transport},
  author  = {Liu, Qiang},
  journal = {arXiv preprint arXiv:2209.14577},
  year    = {2022}
}

@inproceedings{liu2024instaflow,
  title     = {{InstaFlow}: One Step Is Enough for High-Quality Diffusion-Based Text-to-Image Generation},
  author    = {Liu, Xingchao and Zhang, Xiwen and Ma, Jianzhu and Peng, Jian and Liu, Qiang},
  booktitle = {International Conference on Learning Representations},
  year      = {2024}
}

@inproceedings{yang2025rectified,
  title     = {Text-to-Image Rectified Flow as Plug-and-Play Priors},
  author    = {Yang, Xiaofeng and Chen, Cheng and Yang, Xulei and Liu, Fayao and Lin, Guosheng},
  booktitle = {International Conference on Learning Representations},
  year      = {2025}
}
\end{document}